\documentclass[sts,preprint]{imsart}
\usepackage[utf8]{inputenc}
\usepackage[utf8]{inputenc}
\usepackage[T1]{fontenc}
\usepackage{graphicx}
\usepackage{subcaption}
\usepackage{mathtools}
\usepackage{booktabs}
\usepackage{bm}
\usepackage{tcolorbox}
\usepackage{multicol,multirow}
\usepackage{comment}

\usepackage{booktabs}

\usepackage{enumerate}   
\usepackage{amsmath}
\usepackage{multirow}
\usepackage{bbm}
\usepackage{tikz}

\definecolor{mygreen}{rgb}{0,0.6,0}
\usetikzlibrary{positioning,arrows,decorations,decorations.markings,shadows,positioning,arrows.meta,matrix,fit}

\usetikzlibrary{positioning,arrows}
\usepackage{wrapfig}
\DeclareMathOperator*{\argmax}{arg\,max}
\DeclareMathOperator*{\argmin}{arg\,min}

\usepackage{amsmath}
\usepackage{appendix}
\usepackage{amssymb}
\usepackage[cal=boondoxo]{mathalfa}

\usepackage{amsthm}
\usepackage{hyperref}
\usepackage[capitalise]{cleveref}
\Crefname{assumption}{Assumption}{Assumptions}

\theoremstyle{plain}
\newtheorem{theorem}{Theorem}

\theoremstyle{definition}
\newtheorem{assumption}{Assumption}

\usepackage{graphicx}

\def\Holder{{H\"{o}lder}}

\newcommand{\cm}{\mathcal{M}}
\newcommand{\MDP}{\mathrm{MDP}}
\newcommand{\TMDP}{\mathrm{TMDP}}
\newcommand{\NMDP}{\mathrm{NMDP}}

\newcommand{\cj}{\mathcal{j}}
\newcommand{\cJ}{\mathcal{J}}

\newcommand{\ts}{\textstyle}

\newcommand{\op}{\mathrm{o}_{p}}
\newcommand{\E}{\mathbb{E}}

\newcommand{\pa}{\mathrm{\pa}}

\newcommand{\rI}{\mathrm{I}}

\newcommand{\var}{\mathrm{var}}

\newcommand{\rE}{\mathrm{E}}

\newcommand{\rd}{\mathrm{d}}

\newcommand{\prns}[1]{\left(#1\right)}
\newcommand{\braces}[1]{\left\{#1\right\}}
\newcommand{\bracks}[1]{\left[#1\right]}

\newcommand{\abs}[1]{\left|#1\right|}

\newcommand{\epol}{\pi^\mathrm{e}}
\newcommand{\bpol}{\pi^{\mathrm{b}}}

\newcommand{\DRL}{\mathrm{DRL}}

\newcommand{\dr}{\mathrm{dr}}

\newcommand{\Fcal}{\mathcal{F}}

\newcommand{\Mcal}{\mathcal{M}}
\newcommand{\Scal}{\mathcal{S}}
\newcommand{\Acal}{\mathcal{A}}

\newcommand{\Bcal}{\mathcal{B}}

\newcommand{\Wcal}{\mathcal{W}}

\newcommand{\Qcal}{\mathcal{Q}}

\renewcommand{\eqref}[1]{(\ref{#1})}

\newcommand{\RN}[1]{%
  \textup{\uppercase\expandafter{\romannumeral#1}}%
}

\usepackage{color,graphicx,lscape,longtable}

\def\boxit#1{\vbox{\hrule\hbox{\vrule\kern6pt\vbox{\kern6pt#1\kern6pt}\kern6pt\vrule}\hrule}}

\usepackage{xcolor}
\newcount\Comments  
\newcommand{\kibitz}[2]{\ifnum\Comments=1\textcolor{#1}{#2}\fi}

\usepackage{algorithmic}
\usepackage{algorithm}
\usepackage{comment}
\colorlet{MAGENTA}{magenta}
\newcommand{\revise}{\color{black}}

\RequirePackage{amsthm,amsmath,amsfonts,amssymb}
\RequirePackage[authoryear]{natbib}

\startlocaldefs

\endlocaldefs

\begin{document}

\begin{frontmatter}
\title{A Review of Off-Policy Evaluation in Reinforcement Learning}
\runtitle{Off-Policy Evaluation in 
	Reinforcement Learning}

\begin{aug}
\author[A]{\fnms{Masatoshi } \snm{Uehara}\ead[label=e1]{mu223@cornell.edu}}, 
\author[B]{\fnms{Chengchun } \snm{Shi}\ead[label=e2]{c.shi7@lse.ac.uk}},
\author[C]{\fnms{Nathan } \snm{Kallus}\ead[label=e3]{kallus@cornell.edu }}


\address[A]{Masatoshi Uehara is a Ph.D. student in the Department of Computer Science at Cornell University and Cornell Tech, NYC, USA, \printead{e1}.}
\address[B]{Chengchun Shi is an Assistant Professor in the Department of Statistics at London School of Economics and Political Science, London, UK, \printead{e2}.}
\address[C]{Nathan Kallus is an Associate Professor of Operations Research and Information Engineering, Cornell Tech, NYC, USA, \printead{e3}.}

\end{aug}

\begin{abstract}
Reinforcement learning (RL) is one of the most vibrant research frontiers in machine learning and has been recently applied to solve a number of challenging problems. In this paper, we primarily focus on off-policy evaluation (OPE), one of the most fundamental topics in RL. In recent years, a number of OPE methods have been developed in the statistics and computer science literature. We provide a discussion on the efficiency bound of OPE, some of the existing state-of-the-art OPE methods, their statistical properties and some other related research directions that are currently actively explored. 
\end{abstract}

\begin{keyword}
\kwd{Off-policy evaluation}
\kwd{semiparametric methods}
\kwd{causal inference}
\kwd{dynamic treatment regime}
\kwd{offline reinforcement learning}
\kwd{contextual bandits}
\end{keyword}

\end{frontmatter}

\section{INTRODUCTION}\label{sec:intro}
Reinforcement learning \citep[RL,][]{SuttonRichardS1998Rl:a} has been arguably one of the most vibrant research frontiers in machine learning. It is concerned with learning an optimal policy in sequential decision making problems to maximize the long-term reward that the decision maker receives. 
In the recent few years, RL has been successfully applied to solve a number of challenging problems across various domains, including games \citep{silver2016mastering}, robotics \citep{kober2013reinforcement}, ridesharing \citep{xu2018large} and autonomous driving \citep{sallab2017deep}, to mention a few.  
 
In this paper, we focus on off-policy evaluation (OPE), {\revise a fundamental problem in RL}. OPE is concerned with estimating the mean reward of a given decision policy, known as the \emph{evaluation policy}, using historical data generated by a potentially different policy, known as the \emph{behavior policy}. 
{\revise While OPE may be useful 
for \emph{online RL}, where we can experiment with novel policies, it is most crucial for \emph{offline RL}, where we 
only have access to a historical dataset and are not allowed to experiment. Thus,} 
OPE is important in a number of applications, {\revise where experimentation may be
expensive, risky, and/or unethical,} including healthcare \citep{Murphy2001}, recommendation systems \citep{NIPS2011_4321}, education \citep{Mandel2014}, dialog systems \citep{JiangHaoming2021TAEo} and robotics \citep{levine2020offline}. 
{\revise In these settings, any new policy must first} be evaluated offline based on previously collected historical data before {\revise it can be tried} in practice.
{\revise Moreover, reliable quantification of the uncertainty in the evaluation can be crucial. OPE can thus be phrased as a statistical estimation and inference problem.}

In the statistics literature, OPE is closely related to a line of research on evaluating the outcome of patients under a fixed or data-dependent decision policy \citep{wang2012evaluation,zhang2012robust,ZhangBaqun2013Reoo,chakraborty2014inference,matsouaka2014evaluating,luedtke2016statistical,luedtke2017evaluating,wang2018quantile,zhu2019proper,ShiChengchun2020BtCo,wu2020resampling}. There is in particular a growing interest in developing statistical methods for precision medicine \citep[see e.g.,][for an overview]{tsiatis2019dynamic}, which is a medical paradigm that focuses on identifying the most effective treatment based on individual patient information. See also \citet{MurphyS.A.2003Odtr,Robins2004,qian2011performance,zhao2012estimating,lu2013variable,PhillipJ.Schulte2014QAMf,zhang2015using,fan2017concordance,jiang2017estimation,zhu2017greedy,zhang2018interpretable,shi2018maximin,Ertefaie2018,mo2020learning} and the references therein. Formally speaking, a decision policy (also known as a treatment regime), is a function that takes as input a patient's personal information such as their demographic characteristics, genetic information, clinical measurements and outputs the treatment that they shall receive from among the options available. For many chronic diseases such as cancer and diabetes, treatment of patients involves
a series of decisions. Decision policies in these settings are often dynamic, depending on patients' responses measured over time. These are known as dynamic treatment regimes (DTR). The objective of OPE in these settings  is to evaluate the expected outcome of a (dynamic) treatment policy based on historical data collected from randomized clinical trials or observational electronic health records. 

We note that 
the statistics literature typically adopts a counterfactual or potential-outcome framework \citep{rubin05} to describe the estimand of interest and impose assumptions such as 
unconfoundedness and consistency (see Section \ref{sec:bandit} for details) to guarantee that the estimand can be identified from the observed data. 
To the contrary, researchers in the RL community generally do not use this framework, 
starting instead from a structural assumption of working in, for example, a Markov decision process (MDP). 
Since the
objectives are essentially the same, algorithms developed in the RL literature can be easily adapted to evaluate the impact of a given DTR. Similarly, results established in the statistics literature are applicable to more general OPE problems as well. However, as far as we can tell, researchers that belong to one community are often not aware of works published in the other community, 
and vice versa. 
This motivates us to present an overview of the recent literature on OPE from a unified perspective. We remark that in addition to the statistics and computer science literature, OPE is widely studied in the economics literature as well \citep[see e.g.,][]{KitagawaToru2018WSBT,huber2019introduction}.

Recently, there 
have been 
a few reviews of existing statistical learning methods for precision medicine  \citep[see e.g.,][]{Kosorok2015,KosorokMichaelR2019PM}. However, these methods were primarily motivated by applications in finite horizon settings with only a few treatment stages. In this paper, we focus on settings where the horizon (number of treatment stages) is very long and possibly infinite.
These settings are natural for formulating sequential decision making problems in 
robotics, ridesharing, autonomous driving, and have been widely studied in the computer science literature \citep{SuttonRichardS1998Rl:a}. 
The long- and infinite-horizon setting 
has more recently been adopted in the statistics literature 
on learning and evaluating optimal DTRs in mobile health applications \citep{Audrey2018,Liao2020,LuckettDanielJ.2018EDTR,hu2020personalized,LiaoPeng2020BPLi,shi2020statistical}. 
OPE is challenging in these settings. For example, directly applying the stepwise (augmented) inverse propensity score weighted ((A)IPW) estimator \citep[][]{robins99,ZhangBaqun2013Reoo} developed in the short finite horizon settings might be {\revise 
	inapplicable or sub-optimal} 
due to the following reasons:
\begin{itemize}
    \item {\revise First, these stepwise methods 
    are inconsistent 
    with finitely many  trajectories (e.g., patients). 
	Datasets of this type frequently occur in several mobile health applications \citep[see e.g.,][]{marling2018ohiot1dm} where the total number of patients is small while the number of decision points can be very large.}
    \item {\revise Second, these stepwise methods are known to suffer from the curse of horizon in the sense that its mean squared error (MSE) grows exponentially fast with respect to the horizon \citep{Liu2018,gottesman2019guidelines}. As we will explain, the exponential dependence on horizon is unavoidable when the observed data are generated by a non-Markov decision process (NMDP), which is typically the setting that the existing literature on DTR considers, but it can be avoided if we assume additional structure. }
\end{itemize}

{\revise Recent works from the computer science literature \citep[see e.g.,][]{Liu2018,KallusNathan2019EBtC} suggest that if we are willing to assume some additional structures, \emph{i.e.}, Markovianity and time-homogeneity, we can alleviate these challenges. First, Markovianity makes it possible for us to construct efficient estimators to break the curse of horizon. Second, time-homogeneity enables us to construct efficient estimators even from a single trajectory. We remark that most works in computer science consider settings with these structures. These settings differ from those studied in the classical statistics literature.}

This article reviews the OPE problem from a statistical perspective 
and focusing on the Markovian setting with long or infinite horizon. 
We 
review 
what is the best-possible asymptotic MSE (also known as the efficiency bound) that one can hope for in OPE and discuss ways to construct efficient estimators 
that 
achieve this bound. Efficiency bounds have been recognized as 
{\revise natural criteria}
to {\revise understand} the statistical limits of parameter estimation and to compare different estimators \citep{LaanMarkJ.vanDer2003UMfC}. We will also discuss some recent OPE methods developed in the computer science literature, summarize their advantages and limitations, and propose possible directions for future development.

The rest of the article is organized as follows. In \cref{sec:bandit}, for pedagogic purposes, we start with 
\emph{non-dynamic} 
setting 
\citep{DudikMiroslav2014DRPE}, a special case of RL with independent transitions, 
where we 
present the efficiency bound and introduce  various estimators with different statistical properties. 
Section \ref{sec:nmdp_tmdp} is concerned with OPE under an NMDP or time-varying MDP and Section \ref{sec:mdp} under a (time-homogeneous) MDP.
We introduce the efficiency bounds 
for each setting 
and present methodologies to construct efficient estimators that achieve these bounds 
using two key functions, Q-functions and marginal density ratios. 
{\revise We further briefly touch on model-based estimators and compare model-free estimators that employ Q-functions and marginal ratios.  
 In \cref{sec:finite}, we summarize additional important statistical theories of OPE, such as convergence rates of Q-functions and marginal density ratios, and what makes OPE problems special, compared to standard conditional moment problems in statistics.} 
In \cref{sec:extension}, we give an overview of some other related research directions that are currently 
being
actively explored. 

\section{
NON-DYNAMIC 
SETTING}\label{sec:bandit}

To begin with, we consider estimating a given evaluation policy's expected return (value) in the 
non-dynamic 
setting 
(sometimes referred to as the contextual bandit setting, even if not studying adaptive experimentation). 
The problem can be formulated as follows. 
The decision maker 
first observes some contextual information (
also called 
baseline covariates 
or 
state variables), summarized into $S \in \Scal$ in some domain $\Scal$.  An action (
also called 
treatment 
or 
intervention) $A$ is then selected out of a set of candidates $\Acal$ according to some behavior policy $\bpol$ \citep[also known as the propensity score,][]{rosen83} such that $\bpol(\cdot \mid s)$ corresponds to the probability mass or density function of $A$ conditional on $S=s$, depending on whether $A$ is discrete or continuous. In return, the decision maker receives a numerical reward $R \in \mathbb{R}$ for the chosen action. This process repeats for multiple steps, yielding the following observed dataset:
{\revise 
\begin{align}\label{eq:stratified}
  \{\mathcal{J}^{(i)}\}_{i=1}^n= \{S^{(i)},A^{(i)},R^{(i)}\}_{i=1}^{n}\overset{\text{i.i.d}}\sim P_{\bpol}(s,a,r)=p_{S}(s)\bpol(a|s)p_{R|S,A}(r|s,a),
\end{align}
}where $(S^{(i)},A^{(i)},R^{(i)})$ denotes the state-action-reward triplet observed at time $i$ and the notation $P_{\bpol}$ is used to indicate that the data are generated according to the behavior policy. In addition to the observed reward, we use $R(a)$ to denote the potential outcome, representing the reward that the decision maker would have received if action $a$ is selected. We remark that the above setting is essentially the same 
as in 
the statistics literature on causal inference \citep[see e.g.,][]{imbens_rubin_2015,hernan2019} and estimation of optimal individualized treatment regimes in point exposure studies \citep[see e.g.,][]{zhang2012robust}.  

In OPE, our goal is to estimate the expected reward of a given evaluation policy $\epol$. Formally speaking, $\epol$ is a function that maps the space of contextual information $\Scal$ to a probability distribution on $\Acal$. 
When the action space is discrete, the decision maker will set $A=a$ with probability $\epol(a|s)$ under $\epol$. In the causal inference literature, most papers consider settings of binary treatment and state-agnostic evaluation policies, i.e., $\Acal=\{0,1\}$ and that $\epol(a|s)=\rI(a=1)$ or $\epol(a|s)=\rI(a=0)$.  They focus on estimating the average treatment effect (ATE), defined as the value difference between the two state-agnostic policies. When the action space is continuous, we require $\epol$ to be a stochastic policy, i.e., $\epol(\cdot |s)$ corresponds to a non-degenerate probability density function for any $s$. Some extra care is needed when $\epol$ is a deterministic policy, i.e., $A$ is a deterministic function of $S$ under $\epol$. We discuss this in detail in \cref{ite:conti}. 
Given $S$, let $R(\epol)=\sum_{a\in \Acal} R(a)\epol(a|S)$ denotes the potential outcome that would have been observed under $\epol$. We aim to estimate the value $\E[R(\epol)]$ denoted by $J$ based on the dataset in \eqref{eq:stratified}. Toward that end, we need the following assumptions.

\begin{assumption}[Weak positivity]\label{asm:causal3}
The support of $\epol(\cdot  |s)$ is included in the support of $\bpol(\cdot |s)$ for any $s\in \Scal$.  
\end{assumption}

\begin{assumption}[Consistency]\label{asm:causal}
$R=R(A)$, almost surely. 
\end{assumption}
\begin{assumption}[Unconfoundeness]\label{asm:causal2}
{\revise For any $a\in \Acal$, $A$ and $R(a)$ are conditionally independent given $S$. }
\end{assumption}

Assumption \ref{asm:causal3} is slightly weaker than the standard positivity assumption commonly imposed in the literature on learning optimal individualized treatment regimes. 
The consistency assumption 
requires the reward at each time to depend on their own action only. In other words, there is no interference effect across time. Assumption \ref{asm:causal2} is commonly referred to as the no unmeasured confounders assumption. It automatically holds when the datasets are collected from randomized studies where the behavior policy is a constant function of the contextual information. 

For a given function $f(s,a,r)$ of the state-action-reward triplet, we use $\E_{P_{\pi}}[f]$ to denote its expectation assuming the data follows a given policy $\pi$, i.e., 
\begin{eqnarray*}
	\E_{P_{\pi}}[f]=\int f(s,a,r)P_{\pi}(s,a,r)\mathrm{d}(s,a,r)=\int f(s,a,r)p_{S}(s)\pi(a|s)p_{R|S,A}(r|s,a)\mathrm{d}(s,a,r).
\end{eqnarray*}
In addition, let $\E_n[f]$ denote $\E_n[f]\coloneqq 1/n\sum_{i=1}^n f(S^{(i)},A^{(i)},R^{(i)})$ and $\|f\|_2$ denote $\{\E_{P_{\bpol}}[f^2]\}^{1/2}$.  
Under \cref{asm:causal3,asm:causal,asm:causal2}, we can identify the parameter of interest from the observed data. 
\begin{theorem}
Suppose \cref{asm:causal3,asm:causal,asm:causal2} hold. Let $\eta(s,a)=\epol(a\mid s)/\bpol(a\mid s)$.  
\begin{eqnarray}\label{eqn:Thm1}
	\E [R(\epol)]=\E_{P_{\epol}}[r]=\E_{P_{\bpol}}[\eta(s,a)r].
\end{eqnarray}
\end{theorem}



We make a few remarks. First, the consistency and unconfoundedness assumptions ensure that the first equation in \eqref{eqn:Thm1} holds \citep[see e.g.,][Section 13.3]{TsiatisAnastasiosA2006STaM}. It implies that the target parameter can be written as a function of the observed data. The second equation can be easily verified under the weak positivity assumption. Second, as we have commented, most researchers in the RL community do not adopt the potential outcome framework. However, the developed theories and methods are applicable under \cref{asm:causal,asm:causal2}. 



We next present three commonly-used estimators for policy evaluation. The first one is the importance sampling (IS) estimator, defined by 
\begin{eqnarray*}
	\E_{n}\left[\frac{\epol(a|s)}{\hat \pi^b(a|s)}r\right],
\end{eqnarray*} 
where $\hat \pi^b$ denotes some estimator for $\bpol$. 
The above estimator is also referred to as an IPW or Horvitz-Thompson estimator \citep{hernan2019}. A potential limitation of such an estimator is that when $\epol$ differs greatly from $\bpol$, the ratio $\epol/\bpol$ can be very large for some sample values, making the resulting estimator unstable.  
To overcome this limitation, we could either use a self-normalized IS \citep[NIS,][]{mcbook} estimator defined by 
\begin{eqnarray*}
	\E_{n}\left[\frac{\epol(a,s)}{\hat \pi^b(a|s)}r\right]/\E_{n}\left[\frac{\epol(a|s)}{\hat \pi^b(a|s)}\right], 
\end{eqnarray*}  
or a truncated IS estimator \citep{HeckmanJamesJ.1998Maae} by replacing $\hat \pi^b$ with $\max(\epsilon,\hat \pi^b)$ for some small constant $\epsilon>0$. 

The second estimator is the direct method (DM) estimator, defined by $\E_{n}[\hat q(s,\epol)]$ where $q(s,a)=\E[R|S=s,A=a]$, $\hat q$ denotes some estimator of $q$, and $\hat q(s,\epol)$ is a shorthand for $\E_{a \sim \epol(a|s)}[\hat q(s,a)|s]$. Such an estimator is also referred to as a regression-type estimator. {\revise Notice that DM is biased when the regression model is misspecified. To the contrary, IS is unbiased with known $\bpol$, as in randomized studies. However, IS might have a much larger variance than DM when $\epol$ differs considerably from $\bpol$. This represents a bias-variance trade-off. In addition, DM requires a weaker coverage assumption than IS. For instance, suppose that we use a linear model, e.g., $q(s,a)=\langle \theta,\phi(s,a)\rangle $. Then DM requires $\sup_{x}x^{\top}\E_{P_{\pi^e}}[\phi(s,a)\phi(s,a)^{\top}]x/x^{\top}\E_{P_{\pi^b}}[\phi(s,a)\phi(s,a)^{\top}]x<\infty$. This is weaker than $\max_{(a,s)}\pi^e(a\mid s)/\pi^b(a\mid s)<\infty$, the coverage condition required by IS.}


The third estimator combines the first two for more robust policy evaluation. It is referred to as the doubly robust estimator, defined as 
\begin{align*}
   \hat J_{\dr}=\E_n\bracks{\frac{\epol(a|s)}{\hat \pi^b (a|s)}\{r-\hat q(s,a)\}+\hat q(s,\epol)}. 
\end{align*}
Such an estimator has been extensively studied in the statistics literature \citep[see e.g.,][]{RobinsJamesM.1994EoRC,scharfstein99,bang2005doubly,TsiatisAnastasiosA2006STaM,zhang2012robust,ChernozhukovVictor2018Dmlf,FosterDylanJ.2019OSL,chernozhukov2018biased}. \citet{DudikMiroslav2014DRPE} introduced this estimator in the machine learning literature. 

We next briefly discuss some statistical properties of the doubly robust estimator. 
First, it is doubly robust in that as long as either model for $\hat q(s,a)$ or $\hat \pi^b(a|s)$ is well-specified, the final estimator is consistent. 

{\revise Second, it is semiparametrically efficient in that its MSE achieves the efficiency bound under certain rate conditions specified below. Formally speaking, given a model for the joint distribution function of the state-action-reward triplet, the efficiency bound is defined as the lower bound of the asymptotic MSEs among all regular $\sqrt{n}$-consistent estimators with respect to the model \citep[see e.g., ][Chapters 7 and 25]{VaartA.W.vander1998As}. More specifically, regular estimators are those whose limiting distributions are insensitive to small changes to the data generating process (DGP) within the region of the model and the efficient estimator is the one whose MSE is asymptotically equivalent to the efficiency bound. The efficient influence function (EIF) plays a key role in constructing efficient estimators. One popular approach to obtain the EIF is to represent the derivative of the target with respect to (w.r.t.) the model parameters as the integral of a product of a certain function and a score function \citep[see e.g.,][Section 3.4]{kennedy2022semiparametric}. When the model is nonparametric (as required in Theorem \ref{thm:bandit}), this function is unique and equals the EIF. Given the EIF, its empirical average produces the efficient estimator.} 


Theorem \ref{thm:bandit} below summarizes the efficiency bound and the associated EIF of policy value w.r.t. the nonparametric model. Its detailed proof can be found in \citet{RobinsJamesM.1994EoRC,narita2018}. 
\begin{theorem}\label{thm:bandit}
The EIF of $J$ w.r.t. the nonparametric model $\cm$ is 
\begin{align}\label{eq:bandit}
    \eta(s,a)\{r-q(s,a)\}+q(s,\epol)-\E [R(\epol)]. 
\end{align}
The efficiency bound of $J$ w.r.t.  $\cm$ is 
\begin{align*}
V(\cm)=\E_{P_{\bpol}}\bracks{\eta^2(s,a)\var_{P_{\bpol}}[r|s,a]}+\var_{P_{\bpol}}[q(s,\epol)]
\end{align*}
where $\var_{P_{\bpol}}[r|s,a]$ denotes the conditional variance of $R$ given $(A,S)=(a,s)$. 
\end{theorem}
{\revise 
Notice that the model $\cm$ specifies the joint distribution of the state-action-reward triplet. As such, the $q$-function, the behavior policy (e.g., conditional distribution of the action given the state) and the density ratio are determined by $\cm$. 
In addition, as we have commented, an empirical average of the EIF produces the efficient estimator $\E_n [\eta(s,a)\{r-q(s,a)\}+q(s,\epol)]$. However, since $q(s,a)$ and $\eta(s,a)$ are unknown, we need to plug-in their estimators. The resulting value estimator equals $\hat J_{\mathrm{dr}}$.   } 

Third, sample splitting allows us to use one part of the data to learn the nuisance functions $q$ and $\pi^b$, and the remaining part to do the estimation of the main parameter, i.e., the value. When coupled with sample splitting, the scaled MSE $n\E[|\hat J_{\dr}-\E [R(\epol)]|^2]$ achieves the efficiency bound under mild rate conditions on the estimated nuisance functions $\hat q$ and $\hat \pi^b$ \citep{ZhengWenjing2011CTME,ChernozhukovVictor2018Dmlf}. Specifically, we only require $\|\hat \pi^b-\pi^b\|_2=\op(n^{-1/4})$ and $\|\hat q-q\|_2=\op(n^{-1/4})$, allowing these nuisance function estimators to converge at rates that are slower than the usual parametric rate $O_p(n^{-1/2})$ . It allows us to apply the highly adaptive Lasso \citep{BenkeserDavid2016THAL}, random forest \citep{WagerStefan2016ACoR} and deep learning \citep{lecun2015deep} \footnote{{\revise 
The best possible minimax rate that a nonparametric estimator of  $q$ can achieve is given by $n^{-\alpha/(d+2\alpha)}$ under the assumption that $q$ belongs to a Holder ball with dimension $d$ and smoothness parameter $\alpha$ \citep{tsybakov2009lower}. 
It is immediate to see that the rate becomes slower when the state is high-dimensional or the function is non-smooth, leading to the potential violation of the rate condition. We also remark that when $q$ is estimated via state-of-the-art deep learning methods, it is possible to obtain 
convergence rates that depend on the intrinsic dimension instead of the original dimension \citep{schmidt2019deep,nakada2020adaptive,suzuki2018adaptivity}.  In addition, the $o_p(n^{-1/4})$-rate requirement can be further relaxed using the theory of higher-order influence functions \citep{robins2017minimax,mukherjee2017semiparametric}.
} }. 
Without sample splitting, some additional metric entropy conditions need to be imposed on these nuisance estimators and many machine learning estimators might fail to satisfy these conditions \citep{DiazIvan2019Mlit}. The purpose of imposing the metric entropy condition is to verify a stochastic equicontinuity condition in the theoretical analysis. We remark that sample-splitting is commonly used for statistical inference \citep[see e.g.,][]{romano2019multiple}.



Finally, we remark that there exist some specific DM and IS estimators that satisfy the doubly-robustness property or achieve the efficiency bound. 
For instance, \citet{KangJosephD.Y.2007DDRA} developed a DM estimator, which utilizes a weighted regression method to estimate $q$-functions by using weights proportional to $\epol/\bpol$. This estimator is doubly robust. Similarly, a doubly robust IS estimator was proposed by \citet{scharfstein99}. \citet{HahnJinyong1998OtRo} developed a DM estimator that estimates $q(s,a)$ nonparametrically. It achieves the efficiency bound under certain smoothness conditions. \citet{hirano03} proposed an efficient nonparametric IS estimator under similar conditions. 

Although $\hat J_{\dr}$ is widely used, it is known that the vanilla doubly robust estimator $\hat J_{\dr}$ suffers from several limitations, especially when the nuisance function models are misspecified. To address these limitations, a number of estimators have been further developed. First, \citet{RubinDanielB2008Eemi} and \citet{CaoWeihua2009Iear} proposed some alternative doubly robust estimators that achieve minimum asymptotic variance among some class even when the model for $q$ is misspecified.
\citet{TanZhiqiang2010Bead} 
developed a doubly-robust estimator that is bounded and intrinsically efficient, {\revise e.g., guaranteed to be more efficient than IS and NIS estimators asymptotically even when the model for $q$ is misspecified.} All the aforementioned estimators require the correct specification of $\bpol$ to ensure certain optimality properties. 
\citet{VermeulenKarel2015BDRE} proposed a bias-reduced estimator to handle the case where both models are misspecified. There are also some other estimators, including the covariate balancing-type estimator \citep{WangYixin2019MDAB,imai2014covariate,ning2020robust}, the minimax estimator \citep{hirshberg2017augmented,chernozhukov2020adversarial}, the switching estimator \citep{TsiatisAnastasiosA2007CDDR,wang2017optimal}, the target maximum likelihood estimator \citep[TMLE,][]{vanderLaanMarkJ2018TLiD}, the high-order influence function estimator \citep{robins2017minimax,mukherjee2017semiparametric} and the distributionally robust estimator \citep{Nian2020}, among many others \citep[see  e.g.,][]{agarwal2017effective,su2020doubly,sondhi2020balanced,singh2021debiased}.


%
%

\section{OFF-POLICY EVALUATION ON NMDP, TMDP}\label{sec:nmdp_tmdp}
This section is concerned with OPE in general sequential decision making problems with dependent transitions. We focus on two particular data generating processes, NMDP and time-varying Markov decision process (TMDP). As commented in the introduction, estimators in NMDPs suffer from the curse of horizon \citep{Liu2018}. Their MSEs grow exponentially fast with respect to the number of horizon $H$ in general. On the contrary, in TMDPs, there exist estimators whose MSEs grow polynomially in $H$. We will discuss this in detail.  

This section is organized as follows. We first introduce the data generating process and formulate the problem. We next present the DM and IS estimators under these settings and discuss their limitations. Finally, we present the efficiency bounds in NMDP and TMDP, and introduce efficient estimators whose MSEs achieve these lower bounds. 

\subsection{Data Generating Process and Problem Formulation}\label{sec:setup}
We choose not to use the potential outcome framework to simplify the presentation. The data trajectory from NMDP or TMDP can be summarized as a sequence of state-action-reward-next state quadruplets $\cJ=\{(S_t,A_t,R_t,S_{t+1})\}_{1\le t\le H}$ where $H$ denotes the horizon. Similar to the bandit setting, at time $t$, the decision maker observes a state vector $S_t$ from the environment, based on which an action $A_t$ is selected based on the observed data history. The environment responds by providing the decision maker with an immediate reward $R_t \in \mathbb{R}$ and moves to the next state $S_{t+1}$. {\revise Notice that contextual bandits correspond to a special case of NMDPs/TMDPs with $H=1$.}

Let $p_{S_1}$ denote the probability mass/density function of $S_1$. 
Let $\{(s_t,a_t,r_t,s_{t+1})\}_{1\le t\le H}$ denote a realization of $\cJ$. 
In addition, 
let $\cj_{a_t}=(s_1,a_1,r_1,\dots,s_t,a_t)$ and $\cj_{s_t}=(s_1,a_1,r_1,\dots,a_{t-1},s_t)$ denote the state-action-reward history up to $a_t$ and $s_t$, respectively. In an NMDP, we assume the actions are generated according to a history-dependent behavior policy $\pi^b=\{\pi_t^b\}_{1\le t\le H}$ where each $\pi_t^b$ maps $\cj_{s_t}$ to a probability mass/density function $\pi_t^b(\cdot |\cj_{s_t})$ on the action space. In addition, the state transition and the reward distribution are history dependent as well. We use $p_{S_{t+1},R_t}(\cdot |\cj_{a_t})$ to denote the conditional probability mass/density function of $(S_{t+1},R_t)$ given the past data history. We use $P_{\pi^b}$ to denote the joint distribution function of the data trajectory. As commented in the introduction, such a data generating process is commonly used in the literature on learning dynamic treatment regimes. 


A TMDP is a special case of the NMDP where the observed data satisfy the Markov property. Informally speaking, future and past observations are conditionally independent given the present. Specifically,  $p_{S_{t+1},R_t}(\cdot |\cj_{a_t})$ depends on $\cj_{a_t}$ only through the current state-action pair $(s_t,a_t)$. We use $p_{S_{t+1},R_t}(\cdot |s_j,a_j)$ to denote the corresponding conditional distribution. In addition, we assume the behavior policy is a Markov policy. In other words, $\pi_t^b(a_t\mid \cj_{s_t})=\pi_t^b(a_t\mid s_t)$ for any $t$. 

To summarize, the joint distribution of the data from an NMDP can be factorized as follows,
\begin{align}\label{eq:trajdist_nmdp}
P_{\pi^b}=p_{S_1}(s_1)\prod_{t=1}^H \pi_{t}^b(a_t\mid \cj_{s_t})p_{S_{t+1},R_t}(s_{t+1},r_t\mid \cj_{a_t}).
\end{align}
In TMDP, the joint distribution becomes 
\begin{align}\label{eq:trajdist_mdp}
P_{\pi^b}=p_{S_1}(s_1)\prod_{t=1}^H \pi_{t}^b(a_t\mid s_t)p_{S_{t+1},R_t}(s_{t+1},r_t\mid s_t,a_t).
\end{align}
We use $\cm_{\NMDP}$ and $\cm_{\TMDP}$ to denote the nonparametric models specified by \eqref{eq:trajdist_nmdp} and \eqref{eq:trajdist_mdp}, respectively, without any other parametric assumptions. In practice, the underlying data generating process is usually unknown to us. In that case, we can test the Markov assumption based on the observed data to distinguish between NMDP and TMDP \citep{pmlr-v119-shi20c}\footnote{ {\revise When the model selection results are uncertain, e.g., the test sometimes inconsistently classifies a TMDP as an NMDP, the subsequent inference for the off-policy value will be problematic. Problems of this type have been known as ``selective inference" \citep{taylor2015statistical}. In that case, sample-splitting can be employed to account for model selection uncertainty and to ensure valid inference. For instance, we may use one half of the data for model selection and the remaining half for the subsequent inference. Other data-splitting methods can be potentially used as well \citep[see e.g.,][]{meinshausen2009p,luedtke2016statistical,ShiChengchun2020BtCo,shi2021statistical}.} }. 

In OPE,  our goal is to estimate the average cumulative reward of an evaluation policy $\pi^e$,
{\revise 
$
J^{(H)}=\rE_{P_{\pi^e}}\bracks{\sum_{t=1}^H\gamma^{t-1} r_t},
$}
using $n$ data trajectories (e.g., plays of a game or patients in healthcare applications) $\mathcal{D}=\{\cJ^{(1)},\dots,\cJ^{(n)}\}$ that are i.i.d. copies of $\cJ$. 
{\revise Here, a discount factor $\gamma$ ( $0 \leq \gamma \leq 1$)
balances the trade-off between immediate and future rewards.} We require $\pi^e$ to be a Markov policy in TMDP. This assumption is reasonable as there exists an optimal Markov policy whose expected return is no worse than any history-dependent policy under the memoryless assumption 
{\revise \citep{BertsekasDimitriP2012Dpao}.}

To conclude this section, we introduce some additional notations. First, the state-action value functions (better known as the Q-function) and (state) value functions are defined as the conditional expectation of the cumulative reward given the past data history under $\pi^e$. Specifically, we define{\revise
\begin{align*}
q_t(\cj_{a_t})&=\rE_{P_{\pi^e}}\bracks{\sum_{k=t}^H \gamma^{k-t-1} r_k\mid \cj_{a_t}},\,v_t(\cj_{s_t}) =q_t(\cj_{s_t},\pi_t^e)=\rE_{a_t\sim\pi_t^e(a_t\mid \cj_{s_t})}[q_t \mid \cj_{s_t}],
\end{align*}
}for any $t$. {\revise For TMDPs, the data history before a time step $t$ ($\cj_{a_{t-1}}$) is conditionally independent of current and future rewards (e.g., $r_t,\cdots,r_H$) given $s_t,a_t$. With some abuse of notation,  we use $q_t(s_t,a_t)$ and $v_t(s_t)=q_t(s_t,\pi_t^e)$ to represent the corresponding Q- and state value functions under TDMPs. }


%
Second, we denote the density ratio at time $t$ between the evaluation and behavior policy by $
\eta_t(\cj_{a_{t}})=\pi^{e}_{t}(a_t\mid\cj_{s_t})/\pi^{b}_{t}(a_t\mid\cj_{s_t})$. 
We define the cumulative density ratio and the marginal density ratio up to time $t$ as follows, 
\begin{align*}
\lambda_t(\cj_{a_{t}})&=\prod_{k=1}^t \eta_k(\cj_{a_{k}}),\qquad
\mu_t(s_t,a_t)=\frac{p_{\pi^{e}_t}(s_t,a_t)}{p_{\pi^{b}_t}(s_t,a_t)},\quad w_t(s_t)=\frac{p_{\pi^{e}_t}(s_t)}{p_{\pi^{b}_t}(s_t)}, 
\end{align*}
where $p_{\pi_t}(\cdot ,\cdot )$ and  $p_{\pi_t}(\cdot )$ denotes the \emph{marginal} distribution of $S_t,A_t$ and $S_t$, assuming the system follows a given policy $\pi$. 

Finally, for a given function $g(\cj)$ where $\cj=(s_1,a_1,r_1,s_2,a_2,r_2 ,\dots,s_T,a_T,r_H,s_{H+1})$ denotes the entire data trajectory, 
we use $\|g\|_{2}=\rE_{P_{\bpol}}[\abs{g}^{2}]^{1/2}$ to denote its $L^{2}$-norm under the behavior policy. Let $\textstyle\mathrm{E}_{n}[g(\cj)]=n^{-1}\sum_{i=1}^n g(\mathcal \cJ^{(i)}).$ 

\subsection{DM and IS Estimators}\label{sec:litope}

We introduce two popular OPE methods in this section. 
The first approach is the \emph{direct method} (DM), where we estimate the Q-function based on the observed data and directly use it to derive the value estimator. Q-functions are typically estimated by a recursive way \citep{SuttonRichardS1998Rl:a,CliftonJesse2020QTaA}. We will discuss the detailed estimating procedure in \cref{subsec:q_functions}. Once we have an estimate $\hat q_1$ of $q_1$, the DM estimate is given by $$ 
\hat  J^{(H)}_{\mathrm{DM}}=\rE_n\bracks{\hat q_1(s_1,\epol_1)}.$$
One potential limitation of DM is that the estimator might suffer from a large bias due to the model misspecification of Q-function. 

The second approach is the \emph{importance sampling} (IS) method, which averages the data weighted by the density ratio of the evaluation and behavior policies. The resulting estimator is also known as the stepwise IPW estimator \citep{robins99}. Given some estimator $\hat{\lambda}_t$ of the cumulative density ratios $\lambda_t$, 
the IS estimator is defined by 
\begin{align*}
{\revise \hat J^{(H)}_{\mathrm{IS}}=\rE_n\bracks{\sum_{t=1}^H \gamma^{t-1} \hat{\lambda}_t r_t}.}
\end{align*}
When the behavior policy is known and we set $\hat{\lambda}_t$ to its oracle value, the IS estimator is unbiased. However, it might suffer from a large variance due to that $\lambda_t$ can be very large for some sample values. To alleviate this problem, we can consider a self-normalized version of the IS estimator \citep{Precup2000,RobinsJames2007CPoD,KuzborskijIlja2020COEa}, given by
{\revise 
\begin{align*}
    \hat J^{(H)}_{\mathrm{NIS}}=\rE_n \bracks{\sum_{t=1}^H \gamma^{t-1} \braces{\hat{\lambda}_t r_t/\rE_n\bracks{\hat{\lambda}_t}}}.
\end{align*}
}



\subsection{Efficiency Bounds in NMDP and TMDP }\label{sec:finite_bound}

We first introduce the efficiency bound and the EIF for estimating $J^{(H)}$ under NMDP \citep{pmlr-v97-bibaut19a,KallusUehara2019} \footnote{{\revise Special cases of the EIF and efficiency bound have been considered in the literature. In the statistics literature, \citet{RobinsJamesM1995AoSR} studied the case where the evaluation policy is state-agnostic in the context of missing data, and \citet{Murphy2001} considered a deterministic evaluation policy in the context of marginal structural models beyond policy evaluation. In the computer science literature, \citet{jiang} derived the efficiency bound under NMDP with finite state and action spaces. 
However, they did not derive the efficiency bound and the EIF when the state or action space is continuous.
}}. {\revise It extends the contextual bandits results in  \cref{thm:bandit} to sequential decision making.}

\begin{theorem}[Efficiency bound and EIF under $\mathrm{NMDP}$] \label{thm:fin_nnonpara} The efficient influence function of $J^{(H)}$ w.r.t. $\cm_{\NMDP}$ is 
{\revise  
\begin{align}
\label{eq:m1_nonpara}
\phi_{\mathrm{NMDP}}(\cj;\{\lambda_t\},\{q_t\})=-J^{(H)}+
\sum_{t=1}^{H}\{\gamma^{t}\lambda_{t}\prns{r_t-q_{t}}+\gamma^{t-1}\lambda_{t-1}v_t\},\quad v_t= q_t(\cj_{s_t},\pi_t),
\end{align}
}
where {\revise $\lambda_{0}=\gamma$.} The efficiency bound  w.r.t. $\cm_{\NMDP}$  is 
{\revise 
\begin{align}
\label{eq:m1_bound}
V(\cm_{\NMDP})=\mathrm{var}_{P_{\bpol}}[v_0(s_0)] + 
\sum_{t=1}^{H}\mathrm{E}_{P_{\bpol}}\bracks{\gamma^{2(t-1)}\lambda^{2}_{t}(\cj_{a_{t}}) \mathrm{var}\prns{r_{t}+\gamma v_{t+1}\mid\cj_{a_{t}}}}. 
\end{align}
}

\end{theorem}

We next introduce the efficiency bound and the EIF under TMDP \citep{KallusUehara2019} \footnote{{\revise \citet{jiang} derived the efficiency bound under TMDP with finite state and action spaces. 
However, they did not derive the efficiency bound and the EIF when the state or action space is continuous.}}.

\begin{theorem}[Efficiency bound and EIF under $\mathrm{TMDP}$]\label{thm:fin_nnonpara2} The efficient influence function of $J^{(H)}$ w.r.t. $\cm_{\TMDP}$ is 
{\revise 
\begin{align} 
\label{eq:m2_eff}
 \phi_{\mathrm{TMDP}}(\cj;\{\mu_t\},\{q_t\})=-J^{(H)}+
 \sum_{t=1}^{H}\{\gamma^{t}\mu_t\prns{r_t-q_{t}}+\gamma^{t-1}\mu_{t-1}v_t\},\quad  v_t=q_t(s_t,\pi_t), 
\end{align}
}
where {\revise $\mu_{0}=\gamma$. } The efficiency bound  w.r.t. $\cm_{\TMDP}$  is 
{\revise 
\begin{align}
\label{eq:m2_nonpara}
V(\cm_{\TMDP})=\mathrm{var}_{P_{\bpol}}[v_0(s_0)] +\sum_{t=1}^{H}\mathrm{E}_{P_{\bpol}}\bracks{\gamma^{2(t-1)}\mu^{2}_{t}(s_t,a_t)\mathrm{var}\prns{r_{t}+ \gamma v_{t+1}\mid s_t,a_t}}. 
\end{align}
}
\end{theorem}
We make some observations. First, the difference between $\phi_{\mathrm{NMDP}}(\cj;\{\lambda_t\},\{q_t\})$ and $ \phi_{\mathrm{TMDP}}(\cj;\{\mu_t\},\{q_t\})$ is whether $\lambda_t$ or $\mu_t$ is being utilized.
Second, these EIFs and efficiency bounds are derived with respect to {\revise semiparametric models. }
They remain unchanged even if we impose a more restricted model with a known behavior policy. 
Third, the differences between the two bounds characterize the effect of taking into consideration additional problem structures on the feasibility of OPE. 
{
\revise To elaborate, 
assume  (1) $0 \leq R_t \leq R_{\mathrm{max}}$ for all $t=1,\dots,H$, (2) $\mathrm{E}[\log(\eta_t)] \geq \log(C)$ for all $t=1,\dots,H$ and $\E[\log(\mathrm{var}[r_t +\gamma v_{t+1}\mid \cj_{a_t}]) ]\geq \log(V^2_{\min})$,  
(3) $0\leq\mu_t\leq C'$ for all $t=1,\dots,H$. The latter two conditions essentially require that the behavior and evaluation policies do not differ too much. 
Under these assumptions, we can show that
\begin{align*}
V(\cm_{\TMDP})\leq C'R^2_{\max} H^2,\,\,\,\,  V(\cm_{\NMDP})\geq C^{H}V^2_{\min}.
\end{align*}
By the definition of the density ratio, both $C$ and $C'$ are strictly larger than $1$ when the two policies differ from each other. This implies that $V(\cm_{\TMDP})$ grows polynomially fast w.r.t. $H$ if $C'=O(1)$. To the contrary, $V(\cm_{\NMDP})$ grows exponentially fast w.r.t. $H$. } {\revise Notice that $V(\cm_{\NMDP})$ is the smallest possible asymptotic MSE that a regular and asymptotically linear off-policy estimator can achieve in NMDPs, these observations imply that the exponential dependence  on horizon in the estimation error is unavoidable in general without any further assumption. 
} 

{\revise Finally, it is worthwhile to note that the efficiency bound under NMDP can be derived from that under TMDP by embedding NMDPs 
into TMDPs. More concretely, given an NMDP, if we set the state $s^{\diamond}_t$ 
to 
the whole history $\cj_{s_t}=(s_1,a_1\cdots,s_t)$, 
it becomes a TMDP. The resulting efficiency bound is given by
\begin{align}
\label{eq:above_above}
\mathrm{var}_{P_{\bpol}}[v_0(s^{\diamond}_0)] +\sum_{t=1}^{H}\mathrm{E}_{P_{\bpol}}\bracks{\gamma^{t-1}\mu^{2}_{t}(s^{\diamond}_t,a_t)\mathrm{var}\prns{r_{t}+ \gamma v_{t+1}\mid s^{\diamond}_t,a_t}}. 
\end{align}
When the state is history-dependent, the density ratio of the marginal state-action pair is reduced to the cumulative density ratio, i.e., 
$\mu_t(s^{\diamond}_t,a_t)=\lambda_t(\cj_{a_t})$. 
As such, \eqref{eq:above_above} is consistent with the efficiency bound under NMDP displayed in \cref{thm:fin_nnonpara}. 

Despite that the EIFs and efficiency bounds under the two models share similar forms, the inputs of the Q-function and density ratio are different. In particular, both the Q-function and density ratio under NMDP are history-dependent. As such, the resulting EIF suffers from the curse of horizon, yielding a large efficient bound. 
}

\subsection{Efficient Estimators in NMDP}\label{sec:nmdp}

{\revise As commented in Section~\ref{sec:bandit}, a natural idea to obtain an efficient estimator is to take an empirical average of the EIF in \cref{thm:fin_nnonpara} with plugging in estimator for unknown nuisance functions $\lambda_t,q_t$.  }
Given some estimators $\hat{\lambda}_t,\hat q_t$ for $\lambda_t,q_t$, \citet{jiang,thomas2016} proposed the following estimator: 
\begin{align}\label{eq:dr_simplified}
\hat J^{(H)}_{\mathrm{DR}}=\rE_n\bracks{\sum_{t=1}^H\prns{\gamma^t \hat{\lambda}_t(\cj_{a_t})\{r_t-\hat q_t(\cj_{a_t})\}+\gamma^{t-1}\hat{\lambda}_{t-1}\hat q_t(\cj_{s_t},\epol_t)}},
\end{align}
which combines DM and IS for robust policy evaluation. 
When the evaluation policy is state-agnostic or deterministic, similar estimators have been proposed by \citet{RobinsJamesM1995AoSR, Murphy2001,ZhangBaqun2013Reoo}. A split sample version of $\hat J^{(H)}_{\mathrm{DR}}$ was developed by \citet{KallusUehara2019}. 

We next discuss the statistical properties of $\hat J^{(H)}_{\mathrm{DR}}$ when coupled with data splitting. First, it is doubly robust in the sense that the estimator is consistent as long as either the model for $\hat \lambda_t$ or $\hat q_t$ is correctly specified. 
Specifically, when $\{\hat \lambda_t\}_t$ are consistent to $\{\lambda_t\}_t$, we have
\begin{align*}
    \hat J^{(H)}_{\mathrm{DR}}=\E_n\bracks{\sum_{t=1}^H\gamma^{t-1}\hat{\lambda}_{t-1}\hat q_t(\cj_{s_t},\epol_t)-\gamma^t \hat{\lambda}_t\hat q_t(\cj_{a_t})}+\E_n[\sum_{t=1}^H \gamma^t \hat \lambda_t r_t ]\approx  \E_n[\sum_{t=1}^H \gamma^t\hat \lambda_t r_t ]\stackrel{p}{\rightarrow} J^{(H)}. 
\end{align*}
On the other hand, when $\{\hat q_t\}_t$ are consistent to $\{q_t\}_t$, we have
\begin{align*}
\hat J^{(H)}_{\mathrm{DR}}=\rE_n[\hat q_1(s_1,\epol_1) ]+\rE_n\bracks{\sum_{t=1}^H \gamma^t \hat{\lambda}_t\{r_t-\hat q_t(\cj_{a_t})+\gamma \hat q_{t+1}(\cj_{s_{t+1}},\epol_{t+1})\}}\approx \rE_n[\hat q_1(s_1,\epol_1) ]\stackrel{p}{\rightarrow} J^{(H)}.    
\end{align*}
Second, $\hat J^{(H)}_{\mathrm{DR}}$ is efficient under 
{\revise possibly} mild rate conditions on $\{\hat{\lambda}_t\}_t$ and $\{\hat q_t\}_t$. 
Specifically, we only require $\|\hat \lambda_t-\lambda_t\|_2=\op(n^{-1/4})$ and $\|\hat q_t-q_t\|_2=\op(n^{-1/4})$ for any $1\leq t\leq H$ {\revise when coupled with sample splitting.}
{\revise  These rates can be satisfied for many flexible estimators as we will see in \cref{sec:finite}. }
Furthermore, as commented in \cref{sec:finite_bound}, even if this estimator is efficient, the asymptotic MSE grows exponentially fast with respect to $H$. 

Many variants of the DR estimator have been proposed in the computer science and statistics literature, including the self-normalized DR estimator 
\citep{thomas2016}, the more robust DR estimator \citep{Chow2018,TsiatisAnastasiosA2011IDRE}, the TMLE version 
\citep{pmlr-v97-bibaut19a}, the intrinsically efficient estimator \citep{Kallus2019IntrinsicallyES}. We remark that all these methods do not exploit the model structure of TMDPs, i.e., Markovianity. As a result, they \emph{fail} to be efficient under TMDP. 

\subsection{Efficient Estimators in TMDP}

{\revise By taking an empirical average of the EIF in \cref{thm:fin_nnonpara2}
}, \citet{KallusUehara2019} proposed the following double reinforcement learning (DRL) estimator for efficient evaluation under TMDP,
$$
\hat J^{(H)}_{\mathrm{DRL}}=\rE_n\bracks{\sum_{t=1}^H\gamma^t \hat{\mu}_t\{r_t-\hat q_t(s_t,a_t)\}+\gamma^{t-1} \hat{\mu}_{t-1}\hat q_t(s_t,\epol_t)},$$
where $\hat \mu_t$ and $\hat q_t$ denote some estimators for $\mu_t$ and $q_t$, respectively. Similar to $\hat J^{(H)}_{\mathrm{DR}}$, this estimator is consistent when either $\{\hat \mu_t\}_t$ or $\{\hat q_t\}_t$ is consistent, and is efficient when these nuisance function estimators converge at rates $\op(n^{-1/4})$. 
It differs from those estimators outlined in \cref{sec:nmdp} in that its asymptotic MSE grows polynomially in $H$ and is thus much more efficient.  

When $\hat q_t=0$ for any $t$, it is reduced to the marginal IS estimator \citep{XieTengyang2019OOEf,Liu2020_marginal}:
\begin{align*}
    \hat J^{(H)}_{\mathrm{MIS}}=\rE_n\bracks{\sum_{t=1}^H\gamma^t \hat \mu_t r_t }.
\end{align*}
However,  
it is worth mentioning that this estimator is not generally efficient even when $\|\hat \mu_t-\mu_t\|_2=\op(n^{-1/4})$.

Computing $\hat J^{(H)}_{\mathrm{DRL}}$ requires estimating the marginal density ratios and Q-functions from the observed data. We will discuss the estimating procedure of the Q-function in \cref{subsec:q_functions}.
To estimate the marginal density ratio, it suffices to estimate $\eta_t$ (or $\pi_t^b$, equivalently) and $w_t$ since $\mu_t(s_t,a_t)=\eta_t(s_t,a_t)w_t(s_t)$. We note that $\pi_t^b$ can be estimated using existing supervised learning algorithms.
\citet{KallusUehara2019} proposed two approaches to estimate  $w_t(s_t)$. To motivate the first method, we notice that
\begin{align*}
w_{t}(s_t)=\E_{P_{\pi_b}}[\lambda_{t-1}(\cj_{a_{t-1}})\mid s_t].
\end{align*}
Given some estimators for $\lambda_{t-1}$, estimation of $w_t$ can be formulated into a regression problem. To illustrate the second method, we observe that
\begin{align*}
    w_t(s_t)=\E_{P_{\pi_b}}[w_{t-1}(s_{t-1})\eta_t(s_t,a_t)|s_t]. 
\end{align*}
This allows us to sequentially estimate $w_t$ for $t=1,2,\cdots,H$ using regression. 

When parametric models are imposed to learn $w_t$ and $\eta_t$, one can easily show that $\hat \mu_t$ converges at a rate of $O_p(n^{-1/2})$ under mild regularity conditions. {\revise Finally, we remark when behavior policies are unknown, we had better use the following estimating equation:
\begin{align*}
    0 =\E_{P_{\pi_b}}[\mu_{t-1}(s_{t-1},a_{t-1})f(s_t,\pi^e) - \mu_{t}(s_t,a_t)f(s_t,a_t)],\forall f(s,a).  
\end{align*}
The above equation can be used to construct a minimax objective function for estimating $\{\mu_t\}_t$, as discussed in Section~\ref{subsec:estimation_nuisance}. 
}


\section{OFF-POLICY EVALUATION ON MDP}\label{sec:mdp}
This section is concerned with OPE using data generated from a standard Markov decision process \citep[MDP,][]{}.

Similar to TMDP, observations in MDP satisfy the Markov property. However, it differs from TMDP in that the system transitions are homogeneous over time. We remark that most of the existing state-of-the-art RL algorithms rely on this assumption. It ensures the existence of an optimal stationary (time-homogeneous \& Markov) policy that is no worse than any {\revise history-dependent policies 
\citep{puterman2014markov}}. As such, we focus on evaluating a stationary policy $\epol$ throughout this section. 

The rest of this section is organized as follows. We first detail the data generating process and describe the parameter of interest. We next present the efficiency bound and the EIF under MDP. Finally, we review various OPE estimators developed in the literature. 

\subsection{Data Generating Process and Parameter of Interest}
An MDP can be viewed as a special case of the TMDP with time-homogeneous transition functions and infinitely many horizons. Specifically, we have $p_{S_{t+1},R_t}(s_{t+1},r_t|s_t,a_t)=p(s_{t+1},r_t|s_t,a_t)$ for some probability mass/density function $p$ and any $t\ge 0$. Unless otherwise noted, for simplicity, we also assume the behavior policy is stationary, i.e., $\bpol_t=\bpol$ for any $t$. 
It implies that the density ratio is stationary as well, i.e., $\eta_t=\eta$ for any $t$. 
This assumption can be relaxed. Please see e.g., \cite{LiaoPeng2020BPLi} and for further details. 

The offline dataset consists of $n$ trajectories, corresponding to $n$ i.i.d. copies of a data trajectory.  
Let 
$\{(S_t^{(i)},A_t^{(i)},R_t^{(i)},S_{t+1}^{(i)})\}_{1\le t \le H}$ 
be the data collected from the $i$th trajectory where $H$ denotes the number of decision points in the observed data trajectory. 
In addition, the stationarity assumption on the behavior policy does not necessarily imply that $\{S_t\}_t$ is stationary over time. As such, our discussion is applicable to settings where the stationary distribution does not exist, which is certainly the case with a fixed $H$.

The joint distribution of a single data trajectory is given by
\begin{align*}
P_{\pi^b}=p_{S_1}(s_1)\prod_{t=1}^H \pi^b(a_t\mid s_t)p(s_{t+1},r_t\mid a_t,s_t).
\end{align*}
{\revise{The offline dataset can be converted into state-action-reward-next-state tuples $\{S^{[i]},A^{[i]}, R^{[i]},S'^{[i]}\}_{i=1}^{N}$ where
\begin{align}\label{eq:converted}
      S^{[i]}\sim p_b(\cdot),\,\,A^{[i]} \sim \bpol(\cdot|S^{[i]}),\,\,(R^{[i]},S'^{[i]}) \sim p(\cdot,\cdot|A^{[i]}, S^{[i]}),
\end{align}
}}where $N=nH$ and $p_b$ denotes the marginal distribution of the state in the observed data. Hereafter, to simplify the theoretical analysis, we assume 
the data tuples on the left-hand-side are i.i.d. and denote the density function on the right-hand-side as $p_b(s,a,r,s')$. This formulation is standard in offline RL \citep{agarwal2019reinforcement}. For the time-dependent case, we refer the readers to \citet{bibaut2021sequential,shi2020statistical}, which impose certain mixing conditions for the DGPs. 
We use $\cm_{\MDP}$ to denote the nonparametric model specified by the above joint distribution (\ref{eq:converted})  without imposing any additional assumptions.

The parameter of interest we consider is the expected discounted cumulative reward, given by 
{\revise 
\begin{eqnarray}\label{eqn:paraint}
J(\gamma)=\E_{P_{\epol}}\left[\sum_{t=1}^{+\infty} \gamma^{t-1} r_t\right]. 
\end{eqnarray} 
}
{\revise Throughout this section, we suppose $0\leq \gamma<1$. \footnote{\revise{
For mathematical and computational convenience, most existing works in the computer science literature adopt a discounted reward setting where $\gamma$ is strictly smaller than $1$. In the statistics literature, there are some recent works that study OPE under the average reward setting where $\gamma=1$ \citep[see e.g.,][]{liao2021off,LiaoPeng2020BPLi}. RL in these settings involves several technical subtleties arise. For instance, 
the Q-functions need to be defined by subtracting the policy value at each time step to ensure its finiteness. 
Please refer to \citet{LiaoPeng2020BPLi} for more details. 
}}  }
{\revise The quantity $J(\gamma)$  can be interpreted as the evaluation of a functional defined by $\epol(a\mid s),p(s',r\mid s,a)$ and $p_e^{(1)}(s)$, where $p_e^{(1)}(\cdot)$ is an initial state distribution for the target estimand.} For simplicity, we assume $p_e^{(1)}(\cdot)$ is pre-specified.  When $p_e^{(1)}(\cdot)$ is unknown, we can plug in the empirical distribution from the data $\{S^{[i]}_{1}\}_{i=1}^{ N'}\sim p_e^{(1)}(\cdot)$. 

As we will see below, there exist consistent OPE estimators even when the data consist of a single trajectory, i.e., $n=1$. This is different from the findings in Section \ref{sec:nmdp_tmdp} where consistent estimation in NMDPs or TMDPs requires $n$ to diverge to infinity. Such a phenomenon is due to the time-homogeneity assumption under MDP, which enables consistent estimation even with a limited number of trajectories. 


To conclude this section, we introduce the Q-function, the value function and the marginal density ratio under MDP. The Q-function associated with $\pi^e$ is defined as the expected discounted cumulative reward under $\pi^e$ conditional on a given initial state-action pair. Specifically, we have
\begin{eqnarray*}
	q(s,a)=\E_{P_{\pi_e}}\left[\sum_{t=1}^{\infty} \gamma^{t-1} r_t |s_1=s,a_1=a\right].
\end{eqnarray*}
The value function under $\pi^e$ is given by $v(s)=q(s,\epol)=\int \epol(a|s) q(s,a)\rd a$. Finally, we define the marginal density ratio
\begin{eqnarray*}
	\mu^{*}(s,a)=\frac{p^{(\infty)}_{e,\gamma}(s)\epol(a\mid s)}{{p_{b}^{}(s)\bpol(a\mid s)}}, 
\end{eqnarray*}
where 
$p^{(\infty)}_{e,\gamma}$ is the $\gamma$-discounted average state visitation distribution under $\epol$, i.e., $p^{(\infty)}_{e,\gamma}(s)=(1-\gamma) \sum_{t\ge 1} \gamma^{t-1} p_{e}^{(t)}(s)$ where $p_{e}^{(t)}$ denotes the probability mass/density function of $S_{t}$, assuming that the system follows $\pi^e$ and $S_1$ follows $p_{e}^{(1)}$. Here, we implicitly assume that the support of $p^{(\infty)}_{e,\gamma}(s)\epol(a\mid s)$ is included in the support of $p_{b}^{}(s)\bpol(a\mid s)$. This is a weak positivity assumption in MDP. We also define $w^{*}(s)=p^{(\infty)}_{e,\gamma}(s)/p_{b}(s)$ as the marginal density ratio of the state.

\subsection{Efficiency Bound and EIF}

When $p_e^{(1)}$ is pre-specified \footnote{{\revise When $p_e^{(1)}$ is not pre-specified, the efficiency bound will have an additional variance term that is proportional to $\mathrm{var}_{p^{(1)}_e}[v(s) ]$. 
In addition, if we set $\gamma=0$, then the results are reduced to those in Theorem \ref{thm:bandit}. }
}, \cite{KallusNathan2019EBtC} derived the efficiency bound of $J(\gamma)$ w.r.t. a nonparametric model $\cm_{\MDP}$ and established the corresponding EIF. We summarize their results in the following theorem. 
\begin{theorem}\label{thm5}
Letting $v(s')= q(s',\epol)$, the EIF of $J(\gamma)$ w.r.t. $\cm_{\MDP}$ is 
{\revise 
\begin{align}\label{eqn:EIF}
    \phi_{\MDP}(s,a,r,s';\mu^{*},q)=-J(\gamma)+\E_{s\sim p_e^{(1)}}[q(s,\epol)]+(1-\gamma)^{-1}\mu^{*}(s,a)\{r+\gamma v(s')-q(s,a)\},
\end{align}
}
and the efficiency bound $V(\cm_{\MDP})$ is 
{\revise 
\begin{align*}
   V(\cm_{\MDP})=(1-\gamma)^{-2}\E_{p_b}[\mu^{*}(s,a)^2\var[r+\gamma v(s')|s,a]]. 
\end{align*}
}
\end{theorem}


Theorem \ref{thm5} implies that the lower bound of asymptotic MSE among all regular estimators is $V(\cm_{\MDP})/N$. Compared to the lower bound $V(\cm_{\TMDP})/n$ in the TMDP, this rate is faster by a factor of $H=N/n$, which is the benefit we get by taking the time-homogeneity property into consideration.  

{\revise Interestingly, Theorem \ref{thm5} enables us to recover the efficiency bound in TMDPs. Notice that TMDPs can be embedded into MDPs by augmenting the state space 
with a time index $t$. 
Let $\tilde s=(s,t)$ denote the augmented state. It follows that $p_b(\tilde s,a)=H^{-1} p_{\pi^b_t}(s,a)$ and $p^{(\infty)}_{e,\gamma}(\tilde s,a)=(1-\gamma) \gamma^{t-1}p^{(t)}_e(s)$. 
When the initial state distribution under $\epol$ is pre-specified, the variance term $\var_{P_{\pi^b}}[v_0(s_0)]$ does not appear in the efficiency bound and the resulting efficiency bound under this MDP is consistent with that in TMDPs. In addition, due to the inclusion of the time index in the state, the variance of EIF under this MDP is larger than that under a stationary MDP. 


}



We also remark that Theorem \ref{thm5} considers a discounted reward setting. In the literature, \citet{LiaoPeng2020BPLi} derived the efficiency bound for the average-reward  setting where $\gamma=1$, and \citet{bibaut2021sequential} derived the efficiency bound under settings where
{\revise each unit is not independent. }


\subsection{Model-Free Estimators}
We introduce three different types of OPE estimators in this section, corresponding to the DM estimator, the marginal IS estimator and the doubly-robust estimator. All these estimators are model-free in the sense that they are derived without estimating the system transition function $p(s',r\mid s,a)$. In addition, the estimators introduced in Section \ref{sec:nmdp_tmdp} can be potentially applied to estimating 
{\revise $J(\gamma)$.  }
However, as commented before, these estimators are not efficient as they do not take the additional structure into consideration. They require the number of trajectories $n$ to approach infinity to be consistent and converge slower than those estimators we will introduce below. 

First, we introduce the double reinforcement learning (DRL) estimator developed by \cite{KallusNathan2019EBtC}, $$\hat J_{\DRL}=\E_{s\sim p_e^{(1)}}[\hat q(s,\epol)]+\E_N[(1-\gamma)^{-1}\hat \mu^{*}(s,a)\{r+\gamma \hat q(s',\pi^e)-\hat q(s,a)\}],$$
{\revise where the expectation $\E_N$ is taken with respect to the empirical measure $N^{-1} \sum_i \delta_{(S^{[i]}, A^{[i]}, S^{'[i]}, R^{[i]})}$. This DRL estimator is obtained by plugging in certain estimators $\hat \mu^{*},\hat q$ for $\mu^{*},q$ in \eqref{eqn:EIF}.} {\revise We discuss how to construct these estimators in \cref{subsec:estimation_nuisance}. 

Second, we introduce the marginal IS (MIS) and DM estimators. Both estimators can be recovered by DRL with certain choice of nuisance functions. For instance, when set $\hat q=0$, DRL is reduced to the MIS estimator proposed by \citet{Liu2018}, 
{\revise
	\begin{align*}
		\hat J_{\text{MIS}}=  (1-\gamma)^{-1} \E_N[\hat \mu^{*}(s,a)r]. 
	\end{align*}
On the other hand, DRL is reduced to the DM estimator when $\hat \mu^{*}=0$. 
	$$\hat J_{\text{DM}}= \E_{s \sim p_e^{(1)}}[\hat q(s,\epol)].$$
 }As we have mentioned in \cref{sec:bandit}, the DM estimator 
	is more robust to the insufficient coverage of offline data when compared to MIS. More specifically, when we use linear models to parametrize the Q-function, i.e., $q(s,a)=\langle \theta, \phi(s,a) \rangle$, the DM estimator only requires $\sup_{x}x^{\top}\E_{p^{(\infty)}_{e,\gamma}}[\phi(s,a)\phi(s,a)^{\top} ]x/x^{\top}\E_{p_b(s,a)}[\phi(s,a)\phi(s,a)^{\top} ] x <\infty$,   which is weaker than the coverage assumption $\|\mu^{\star}(\cdot)\|_{\infty}<\infty$ required by MIS. In addition, it remains very challenging to consistently estimate the marginal density ratio $\mu^*$. Notice that in contrast to the bandit setting, $\mu^*$ remains unknown even if the behavior policy is known.
	
   Third, when coupled with sample splitting, DRL is semiparametrically efficient when both $\hat \mu^*$ and $\hat q$ converge at a rate of $o_p(N^{-1/4})$. We also remark that when $\hat q$ is parametrized via certain nonparametric estimators such as reproducing kernel Hilbert spaces \citep[RKHSs,][]{steinwart2008support} or linear sieves, the resulting DM estimator is able to achieve the efficiency bound as well. See Theorem 2 of \cite{liao2021off} and Appendix E.2.1 of \cite{shi2022dynamic}, respectively. Similarly, when linear sieves are used to parametrize $\mu^*$, the resulting MIS estimator is semiparametrically efficient as well since it equals DRL with both nuisance functions parametrized via linear sieves \citep{UeharaMasatoshi2019MWaQ}. Nonetheless, different from DRL, all these efficient estimators are nuisance-function-dependent. 
	
	Fourth, DRL estimator is doubly robust in that it is consistent when either $\hat \mu^{*}$ or $\hat {q}$ is consistent. 
	The doubly robust property is confirmed as follows. When 
	$\hat q$ is consistent to $q$, we have 
 {
 	\revise
	\begin{align*}
		\hat J_{\DRL}\approx \E_{s\sim p_e^{(1)}}[\hat q(s,\epol)]\stackrel{p}{\rightarrow} J(\gamma). 
	\end{align*}
 }
	When $\hat \mu^{\star}$ is consistent to $\mu$, we have
	\begin{align*}
		\hat J_{\DRL}\approx (1-\gamma)^{-1}\E_N[\hat \mu^{*}(s,a)r] \stackrel{p}{\rightarrow} J(\gamma). 
	\end{align*}

Finally, it is worthwhile to mention two closely related estimators. First, 
\citet{tang2019harnessing} developed another doubly-robust estimator that breaks the curse of horizon. When the behavior policy is known, the estimator requires either the marginal state density ratio $w^*$ (see the definition in Section \ref{subsecmarginalratio}) or the state value function to be correctly specified. However, its asymptotic variance is larger than DRL; hence, it is generally not efficient \citep{KallusNathan2020EEoN}. Secondly, \citet{shi2021deeply} developed a deeply-debiased estimator for confidence interval construction and uncertainty quantification. It shares similar spirits to the minimax optimal estimating procedure that uses higher-order influence functions
to learn the average treatment effect in contextual bandits  \citep[see e.g.,][]{mukherjee2017semiparametric}. Debiasing brings additional flexibility in that it allows the nuisance functions to diverge at an arbitrary rate. }

\subsection{Estimation of Q-functions $q(s,a)$ and Marginal Ratios $\mu^{\star}(s,a)$ } \label{subsec:estimation_nuisance}

\subsubsection{Estimation of Q-functions $q(s,a)$}

The first method is fitted Q-iteration 
\citep[FQE,][]{ernst2005tree,munos2008finite,FanJianqing2019ATAo}. This is essentially a value iteration method that allows for flexible functional approximation. It recursively updates the Q-estimator based on the Bellman equation
\begin{eqnarray}\label{eqn:bell}
	\E_{(s',r)\sim p(\cdot |s,a)} [r+ \gamma q(s',\pi_e)\mid s,a]=q(s,a),
\end{eqnarray}
for any $(a,s)$. 
More specifically, at the $k$th iteration, it updates $\hat q$ by solving
\begin{align*}
    \hat q_{(k+1)}\leftarrow \argmin_{\tilde  q\in \Qcal}\E_N[\{r-\tilde q(s,a)+\gamma \hat q_{(k)}(s',\epol)\}^2], 
\end{align*}
where $\Qcal$ denotes some flexible function class such as deep neural networks or RKHSs. During each iteration, the above optimization can be cast into a supervised learning problem with $\{R^{[i]}+\gamma \hat q_{(k)}(S^{'[i]},\epol)\}_i$ as the responses, and $\{(A^{[i]}, S^{[i]})\}_i$ as the predictors. 


The second method is minimax Q-learning \citep{UeharaMasatoshi2019MWaQ}. The following observation forms the basis of the method: based on the Bellman equation \eqref{eqn:bell}, we have for any discriminator function $f$ that
\begin{align}\label{eq:estimating}
    \E_{p_b}[f(s,a)\{r+\gamma q(s',\epol)-q(s,a)\}]=0. 
\end{align}
As such, the Q-estimator can be computed by solving the following minimax problem,
\begin{align}\label{eq:minimax}
     \argmin_{\tilde q\in \Qcal}\max_{f\in \Fcal}\E_N[f(s,a)\{r+\gamma \tilde q(s',\epol)-\tilde q(s,a)\}]-\lambda \E_N[f^2(s,a)],\, 
\end{align}
for some tuning parameter $\lambda\ge 0$ and function classes $\Qcal$, $\Fcal$. In view of \eqref{eq:estimating}, the discriminator class $\Fcal$ is introduced to measure the discrepancy between $q$ and $\tilde q$. 
When $\lambda>0$, 
the objective function in \eqref{eq:minimax} corresponds to the 
modified Bellman residual minimization loss \citep[BRM,][]{antos2008learning,Farahm2016,LiaoPeng2020BPLi}. The above minimax optimization is difficult to solve in general. To simplify the calculation, we can set $\Fcal$ to a ball of an RKHS, with which  the inner maximization has a closed-form solution, and then $\hat q$ can be learned by solving the outer minimization via stochastic gradient descent \citep{Liu2018}. Similar to FQE, we can take $\Qcal$ to be a rich function class (e.g., neural networks). Alternatively, we can set both $\Qcal$ and $\Fcal$ to linear models, with which the q-estimator has a closed-form solution. In that case, 
interestingly, the resulting DM estimator $\hat J_{\text{DM}}$ is reduced to LSTDQ {\revise \citep[short for Least-Squares Temporal-Difference learning for Q-functions, see e.g.,][for details]{LagoudakisMichail2004LPI}.} 

{\revise 
Next, we compare minimax Q-learning against FQE. First, from a theoretical point of view, the Bellman closedness assumption required by FQE does not satisfy the monotonic property. In other words, $\Bcal \Qcal\subset \Qcal$ holds does not necessarily imply $\Bcal \widetilde{\Qcal} \subset \widetilde{\Qcal}$ for any $\Qcal\subset \widetilde{\Qcal}$ where $\Bcal$ is a Bellman operator that satisfies $\Bcal \tilde q(\cdot)\coloneqq \E_{p(s',r|\cdot)}[r+\gamma \tilde q(s',\epol)|\cdot] $ for any $\tilde q$. On the contrary, the Bellman closedness assumption required by minimax Q-learning satisfies this property. See \cref{subsec:q_functions} for details. 
Second, from an optimization point of view, 
minimax Q-learning (which relies on minimax optimization) is much more challenging to implement compared to FQE. Finally, it is generally difficult to select hyperparameters in both minimax Q-learning and FQE. In contrast to supervised learning, there does not exist a natural cross-validation criterion in Q-function estimation. Specifically, whereas FQE is built upon supervised learning, the iterative use of it makes cross-validation non-trivial \footnote{ {\revise Recently, this problem has been partially addressed by \citet{miyaguchi2022almost}.}  }. In minimax Q-learning, the minimax objective function makes cross-validation very difficult to implement.}

{\revise Finally, another closely related topic is the estimation of the state-value function $v(s)$.  \citet{FengYihao2019AKLf} proposed a minimax learning method to learn $v$. 
They compute the value estimator by solving a minimax objective function, based on the Bellman equation for the state-value function,
\begin{align*}
    \E_{p_b}[f(s)\eta(s,a)\{r+\gamma v(s',\epol)-v(s)\}]=0, 
\end{align*}
for a certain class of discriminator functions $f\in \Fcal$. 
 In cases where linear models are imposed to model $v$, the resulting method is reduced to those discussed in 
\citet{BertsekasDimitriP2009Pemf,Ueno2011,JMLR:v15:dann14a,LuckettDanielJ.2018EDTR}. }

\subsubsection{Estimation of Marginal Ratios $\mu^{*}(s,a)$}\label{subsecmarginalratio}

We next discuss how to estimate $\mu^{*}(s,a)$. 

\citet{UeharaMasatoshi2019MWaQ} proposed a minimax weight-learning to estimate $\mu^*$ without imposing the stationarity assumption. The method is based on the following identity:
\begin{align}\label{eq:weight_estimating2}
   0= \E_{p_b}[\gamma \mu^{*}(s,a)f(s',\epol)-\mu^{*}(s,a)f(s,a)]+(1-\gamma)\E_{p^{(1)}_e}[f(s,\epol)],\,\forall f(s,a). 
\end{align}
Notice that this equation is agnostic to the behavior policy. 
Then, given certain rich function classes $\Mcal$ and $\Fcal$, the minimax estimator is defined by 
\begin{align*}\ts
       \argmin_{\tilde  \mu^{*}\in \Mcal}\max_{f\in \Fcal} &\E_{N}[\gamma \tilde  \mu^{*}(s,a)f(s',\epol)-\tilde  \mu^{*}(s,a)f(s,a)]+(1-\gamma)\E_{p^{(1)}_e}[f(s,\epol)] -\lambda \E_N[f(s,a)^2],\, 
\end{align*}
for some tuning parameter $\lambda\ge 0$. Interestingly, when we set both $\Mcal$ and $\Fcal$ to linear models, the resulting marginal IS estimator $J_{\text{MIS}}$ is reduced to LSTDQ.

It is worthwhile to add two remarks. First, recalling  $\mu^*(s,a)=w^*(s)\epol(a\mid s)/\bpol(a\mid s)$, when we know $\bpol(a\mid s)$, it suffices to estimate $w^*$. \citet{Liu2018} \footnote{{\revise Notice that \eqref{eq:weight_estimating1} is slightly different from the original proposal in \citet[Lemma 3]{Liu2018} since they take the expectation with respect to the discounted occupancy distribution $p_{b,\gamma}(s)$. Later, the modified version \eqref{eq:weight_estimating1} is discussed in \citet[Lemma 16]{KallusNathan2019EBtC}.} } propose the following estimating equation:
\begin{align}\label{eq:weight_estimating1}
    0=\E_{p_b}[\gamma w^{*}(s)\eta(s,a)f(s')-w^{*}(s)f(s)]+(1-\gamma)\E_{p^{(1)}_e}[f(s)],\,\forall f(s),
\end{align}
where $\eta(s,a)=\epol(a\mid s)/\bpol(a\mid s)$.
{\revise Second, several related minimax estimators have been proposed from a duality viewpoint \citep{YangMengjiao2020OEvt,NachumOfir2020RLvF,dai2020coindice}. More specifically, they cast an OPE problem into a  linear programming problem. Then, the objective function of minimax estimators is derived from the corresponding Lagrange function.   }

\subsection{Model-Based Estimators}
So far we have discussed the model-free method. 
In this section, we focus on the class of model-based estimators. A typical model-based estimator estimates the transition density and reward density functions from the data and plugs in these estimators in \eqref{eqn:paraint} to construct the value estimator. 
Note this estimator is also known as g-formula in the literature on causal inference \citep{hernan2019}. 
When probabilistic neural networks are used to model the transition function, the resulting model-based estimator has shown good empirical performance in challenging continuous control domains \citep{ZhangMichaelR2021ADMf}. 

Statistical properties of model-based estimators have been recently established in the computer science literature. 
\citet{Yin2020} proved that the model-based estimator in discrete MDPs is asymptotically efficient. 
This result extends the findings of \citet{HahnJinyong1998OtRo} to sequential decision making. 
In continuous domains, \citet{uehara2021pessimistic} studied model-based estimators with maximum likelihood parameter estimation of transition density. 

{\revise
To conclude this section, we discuss the advantages and disadvantages of these model-based estimators when compared against model-free estimators. On one hand, hyperparameter selection is much easier in model-based methods since existing state-of-the-art supervised learning algorithms are applicable to learn the state transition and reward functions, and cross-validation can be potentially employed for parameter tuning. On the contrary, as commented in 
\cref{subsec:estimation_nuisance}, hyperparameter tuning is more delicate in model-free methods. On other hand, in settings with high-dimensional state information, model-based estimators might not be preferable since it is more challenging to model the state transition function than to model the value function. 
 }

{\revise 
\subsection{OPE and Nonparametric Instrumental Variable Estimation}

 According to \eqref{eqn:bell}, Q-functions are characterized as solutions to certain conditional moment equations. As such, Q-function 
 estimation (and the subsequent policy evaluation) can be cast into nonparametric instrumental variable estimation, 
 which has long been studied 
 in statistics and econometrics \citep[see e.g.,][]{chamberlain1992comment,ai2003efficient,ai2012semiparametric,newey2013nonparametric,DikkalaNishanth2020MEoC}. This connection has been widely recognized among RL researchers \citep{KallusNathan2019EBtC,chen2022well,zhang2022off}. 
 
 In the rest of this section, we highlight two key features of OPE. First, while Q-function estimation can be formulated into instrumental variable estimation, the Bellman equation \eqref{eqn:bell} has a unique structure where the Q-function appears on both the left-hand-side and right-hand-side of the estimating equation with different inputs. This special structure gives license to use sequential regression, i.e., FQE for Q-function estimation. However, to our knowledge, 
 sequential regression has not been applied to solving standard conditional moment equations. Second, in OPE, the ultimate goal is 
 to estimate the target policy's value. There exist consistent OPE estimators such as $\hat J_{\mathrm{MIS}}$ that can be constructed without Q-function estimation. 
 }

\section{OPE Theory}\label{sec:finite}

{\revise In \cref{sec:mdp}, we mainly discuss the efficiency, assuming that we obtain convergence rates of Q-functions and marginal ratios. In this section, we mainly explain the convergence properties of these two functions and the special characteristics of OPE problems. 
In this section, for simplicity, while we focus on MDPs, every discussion is easily applicable to TMDPs. }

\subsection{Convergence Rates of Q-functions}\label{subsec:q_functions}

We discuss the statistical properties of Q-functions.  
To establish the rate of convergence properties of the aforementioned algorithms,
we often require the realizability and completeness assumptions. Both assumptions are commonly imposed in the computer science literature \citep[see e.g.,][]{munos2008finite,ChenJinglin2019ICiB}. The realizability assumption essentially requires $q\in \Qcal$. In other words, the function class $\Qcal$ shall be rich enough to contain $q$. The completeness assumption requires $\Qcal$ to be closed under the Bellman operator. 
For FQE, the completeness assumption requires $\Bcal \Qcal\subset \Qcal$ (recall that $\Bcal$ is the Bellman operator). It is satisfied when the transition density $p$ is a smooth function and $\Qcal$ contains the class of smooth functions \citep{munos2008finite,FanJianqing2019ATAo}. {\revise  Compared to the realizability, this $\Bcal \Qcal\subset \Qcal$ does not have a monotonic property in that the larger $\Qcal$ does not result in the weaker assumption. For minimax Q-learning, the completeness assumption requires $\Bcal\Qcal\subset  \Fcal$. Similarly, with a smooth transition density function, it holds when $\Fcal$ contains the class of smooth functions. Compared to $\Bcal \Qcal\subset \Qcal$, this $\Bcal\Qcal\subset  \Fcal$ has a certain monotonic property in that the larger $\Fcal$ result in the weaker assumption.}

{\revise Under these conditions $q\in \Qcal$ and Bellman completeness ($\Bcal \Qcal \subset \Qcal$ in FQE and $\Bcal \Qcal \subset \Fcal$ in minimax Q-learning)}, we can derive the 
rate of convergence of the Bellman residual error $\{\E_{p_b}[\{\Bcal \hat q-\hat q\}^2]\}^{1/2}$ as a function of the critical radii of the function class \citep{UeharaMasatoshi2021FSAo,DuanYaqi2021RBaR}. For example, when  we use 
parametric models with finite VC dimensions for $\Qcal$ and $\Fcal$, the rate of convergence is $O_p(N^{-1/2})$. When we use \Holder\,classes with an input dimension $d$ and the smoothness parameter $\alpha$, the rate is $O_p(N^{-\alpha/(2\alpha+d)})$. 
These results can be directly translated into an error bound for $|\hat J_{\text{DM}}-J|$ {\revise using  $$|\hat J_{\text{DM}}-J|\leq \{\E_{p_b}[\mu^{\star}(s,a)^2 ]\}^{1/2}\E_{p_b}[\{\Bcal \hat q-\hat q\}^2(s,a)]\}^{1/2}.$$ 
For the derivation, please refer to \cite{UeharaMasatoshi2021FSAo}. Finally, we remark that Bellman residual errors can be further translated into $\ell_2$-errors $\{\E_{p_b}[(\hat q-q)^2(s,a)]\}^{1/2}$ under certain mild conditions \citep{chen2022well}. Furthermore, \citet{huang2022beyond} discuss how to directly obtain $\{\E_{p_b}[(\hat q-q)^2(s,a)]\}^{1/2}$ under the realizability $q\in \Qcal$ and $q'\in \Wcal$ where $q'$ is a certain function without going through Bellman residual errors.  }

\subsection{Convergence Rates of Marginal Ratios $\mu^{*}(s,a)$}


Finally, we discuss the statistical properties of the marginal IS estimator. We need the realizability condition $\mu^*\in \Mcal$ and the completeness condition $\Bcal'\Mcal\subset \Fcal$ where $\Bcal'$ denotes the adjoint Bellman operator \citep[see][for the detailed definition]{UeharaMasatoshi2021FSAo}. Similar to the Bellman operator $\Bcal$ that satisfies $\Bcal q=q$, $\Bcal'$ satisfies the identity that $\Bcal' \mu^*=\mu^*$ and can be understood as the analog of $\Bcal$ for describing the marginal density ratio. 
Under these conditions, we can characterize the Bellman residual error $\{\E_{p_b}[\{\Bcal'\hat \mu^{*}-\hat \mu^{*}\}^2]\}^{1/2}$ using the critical radii of the function class. 
For instance, when we use parametric models with finite VC dimensions for $\Mcal$ and $\Fcal$, the Bellman residual error decays to zero at a rate of $O_p(N^{-1/2})$. When we use \Holder\,classes with an input dimension $d$ and the smoothness parameter $\alpha$, the rate is $O_p(N^{-\alpha/(2\alpha+d)})$. Similarly, these results can be directly translated into an error bound for $|\hat J_{\text{MIS}}-J|$. Please refer to \cite{UeharaMasatoshi2021FSAo} for details. {\revise 
Furthermore, \citet{huang2022beyond} discuss how to directly obtain $\{\E_{p_b}[(\hat \mu^{\star}-\mu^{\star})^2(s,a)]\}^{1/2}$ under the realizability $\mu^{\star}\in \Mcal$ and ${\mu'}^{\star}\in \Fcal$ where ${\mu'}^{\star} \in \Fcal$ is a certain function without going through Bellman residual errors.
}


{\revise \subsection{Do We Really Need Bellman Completeness?} \label{subsec:dowe}

We discuss the role of the completeness assumption in this section. 
Several papers \citep[see e.g.,][]{du2019good,foster2021offline} proved that the realizability 
condition alone is insufficient to obtain a non-asymptotic error bound that is polynomial in the number of horizon in a variety of contexts. These results suggest that some other conditions are needed in addition to 
the realizability of Q-functions. 

One example of these assumptions is given by the Bellman completeness 
condition introduced in \cref{subsec:q_functions}. Alternatively, we can impose certain matrix invertibility conditions when specialized to linear models \citep{perdomo2022sharp}. These conditions have also been introduced in the statistics literature \citep[see e.g.,][]{Ertefaie2018,shi2020statistical} to study the rate of convergence and asymptotic normality of the Q-function estimator. However, 
it remains unclear how to extend this condition to general function approximation that permits the use of deep neural networks and random forests. 
Finally, \citet{UeharaMasatoshi2021FSAo} obtained a favorable non-asymptotic result 
that allows general function approximation by assuming the realizability of the marginal density ratios. Notably, they proved the convergence of the policy value without the convergence of nuisance function estimators.
}

\section{Recent progress and Discussion}\label{sec:extension}

In this section, we present an overview of some other related research directions that are currently being actively explored. 
\subsection{Non-asymptotic Lower bounds}
In this paper, we present the efficiency bound (e.g., the best-possible asymptotic MSE) for OPE under various DGPs. In the RL community, researchers are particularly interested in non-asymptotic results. The non-asymptotic minimax (or locally minimax) lower bounds characterize the best-possible nonasymptotic MSE as a function of the sample size, the horizon as well as some other quantities of interest. 
These non-asymptotic lower bounds for OPE have been recently established under some specific settings  \citep{duan2020minimax,YinMing2020NOPU,Wang2020,PananjadyAshwin2021,hao2021sparse,mou2021optimal,mou2022off}.

\subsection{Unmeasured confounding}
In observational data, the sequential ignorability assumption could be violated in some applications \citep[see e.g.,][]{NamkoongHongseok2020OPEF}. In that case, the policy value is not identified without additional assumptions. To address this problem, several approaches have been developed in the literature. 
The first line of research proposes to develop partial identification bounds for the policy value based on a sensitivity model \citep{Kallus2020_confounding,NamkoongHongseok2020OPEF,Zhang2021}. These papers are inspired by works in the causal inference literature on sensitivity analysis  \citep{ManskiCharlesF1995Ipit,RosenbaumPaulR.2002OS}. 
{\revise The second line of research focuses on the confounded MDP model \citep{zhang2016markov} under which the Markov assumption is satisfied and relies on certain proxy variables for consistent OPE \citep[see e.g.,][]{bennett2021off,fu2022offline,shi2022off}. The third line of research 
uses partially observable environments to formulate the OPE problem under unmeasured confounding.} For instance, in the discrete state-action space setting, \citet{Tennenholtz_Shalit_Mannor_2020} uses an idea of negative controls \citep{miao2018identifying} and establishes the point identification strategy. {\revise Later, these results are extended to the continuous state-action space setting \citep{bennett2021,miao2022off,shi2022minimax}. 
 }
However, they all rely on certain assumptions that might be difficult to validate in practice. For instance, to derive the partial identification bound, \citet{NamkoongHongseok2020OPEF} assumed that there is only a single step of confounding at a known time step {\revise to sharpen the partial identification bound}. \citet{Tennenholtz_Shalit_Mannor_2020} imposed certain matrix invertibility assumptions to guarantee the identifiability of the policy value from the observed data.

\subsection{Partially Observable MDPs}

{\revise In partially observable MDPs (POMDPs), agents only have access to noisy observations instead of the underlying states. Compared to MDPs, POMDPs are more general. In addition, POMDPs might fit real datasets better than MDPs since the Markovian assumption are often questionable in practice. Since POMDPs are NMDPs, we can employ any off-the-shelf OPE methods for NMDPs \citep{futoma2020popcorn,hu2021off}. However, these methods suffer from the curse of horizon and cannot consistently learn the policy value from a single trajectory. To alleviate these issues, by fully leveraging structures of POMDPs, \citet{uehara2022future} recently proposed a model-free OPE method in POMDPs. 
However, it remains unclear whether there are model-based methods that 
can circumvent the curse of horizon and learn from a single trajectory.
} 

\subsection{Offline policy optimization}
In offline policy optimization, we aim to identify an optimal policy based on the observed data to maximize the expected return. 
A simple idea would be to first construct some OPE estimator $\hat J(\pi)$ 
for any policy $\pi$ that belongs to a policy class $\Pi$, and then compute the optimal policy by $\hat \pi=\argmax_{\pi \in \Pi}\hat J(\pi)$. Such an idea has been adopted in the contextual bandit setting for offline {\revise policy optimization  \citep{zhang2012robust,ZhaoYingqi2012EITR,Swaminathan2015b,swaminathan15a,AtheySusan2017EPL,KitagawaToru2018WSBT}. }However, it is challenging to extend these methods to the long horizon or infinite horizon settings in RL. First, the nuisance functions such as the Q-function and the marginal density ratio involved in the OPE estimator are policy-dependent. It remains unknown how to learn the set of nuisance functions over a given policy class in a computationally efficient manner. 
Second, 
a crucial challenge of applying such a method lies in the existence of out-of-distribution actions due to the mismatch between the behavior policy and the target policy. It results in overestimation of the value evaluated at these out-of-distribution actions, therefore worsening the performance of policy learning. To address this limitation, we can employ the pessimism principle to prevent overestimation and restrict the learned policies to stay close to the behavior policy 
\citep{Swaminathan2015b,swaminathan15a,Yu2020,Kidambi2020}. 
Such a principle has been recently investigated from a theoretical perspective \citep{JinYing2020IPPE,RashidinejadParia2021BORL,uehara2021pessimistic,xie2021bellman}. 

\subsection{General policies}\label{ite:conti}
So far we have focused on the setting where the evaluation policy is known to us. 
There are applications where the evaluation policy is a deterministic function of behavior policy, such as the tilting policy or the modified treatment policy \citep{DiazIvan2019Mlit,young2014identification}. Since the behavior policy is unknown, so is the evaluation policy. The resulting efficiency bound for OPE and the EIF are different from what we have discussed in the main text \citep{KennedyEdwardH2019NCEB,KallusNathan2020EEoN}. 

In addition, as we have discussed in the main text, some extra care is needed when the action space is continuous and the evaluation policy is deterministic. In that case, the causal estimand is no longer pathwise differentiable and $\sqrt{n}$-consistent estimation does not exist \citep{BibautAurelien2017Dsfo,kennedy2017non}. By imposing certain smoothness conditions over the action space, the policy value can still be consistently estimated if we replace the indicator function in the estimating function with kernels \citep{KallusNathan2020DROV}. However, such a method might suffer from the curse of dimensionality and perform poorly in high-dimensional action space.


\subsection{General causal graphs}
NMDP and TMDP are special cases of causal directed acyclic graphs (DAGs). 
For general causal graphs, under the no unmeasured confounders assumption, it is well-known that the policy value is identifiable and that the EIF exists \citep{LaanMarkJ.vanDer2003UMfC,RotnitzkyAndrea2020EASf}. When there are unmeasured variables, 
although a general identification condition exists
\citep{TianJin2002Agic,Shpitser2006}, the estimating procedure is less explored. In some special cases, several estimators have been proposed and are proven to achieve efficiency bound \citep{FulcherIsabelR2020Riop,bhattacharya2020semiparametric,Jung2020,smucler2022efficient}. However, it remains unknown whether nonparametrically efficient estimators exist under more general settings. 

\bibliographystyle{imsart-nameyear}
\bibliography{rc}

\end{document}